# A Stabilized Feedback Episodic Memory (SF-EM) and Home Service Provision Framework for Robot and IoT Collaboration

Ue-Hwan Kim and Jong-Hwan Kim, *Fellow, IEEE*

*Abstract*—The automated home referred to as Smart Home is expected to offer fully customized services to its residents, reducing the amount of home labor, thus improving human beings' welfare. Service robots and Internet of Things (IoT) play the key roles in the development of Smart Home. The service provision with these two main components in a Smart Home environment requires: 1) learning and reasoning algorithms and 2) the integration of robot and IoT systems. Conventional computational intelligence-based learning and reasoning algorithms do not successfully manage dynamic changes in the Smart Home data, and the simple integrations fail to fully draw the synergies from the collaboration of the two systems. To tackle these limitations, we propose: 1) a stabilized memory network with a feedback mechanism which can learn user behaviors in an incremental manner and 2) a robot-IoT service provision framework for a Smart Home which utilizes the proposed memory architecture as a learning and reasoning module and exploits synergies between the robot and IoT systems. We conduct a set of comprehensive experiments under various conditions to verify the performance of the proposed memory architecture and the service provision framework and analyze the experiment results.

## I. Introduction

TECHNOLOGICAL advances in communication, mobile computing, artificial intelligence, and robotics are transforming various aspects of human society, including industry [1], finance [2], education [3], and life-style [4], [5]. Automation of factories leads to increased productivity, people can conduct financial transactions safely, and students can take classes of prestigious professors anywhere in the world at anytime. Amid these changes, even the private

Manuscript received May 11, 2018; revised September 21, 2018; accepted November 12, 2018. This work was supported by the Institute for Information and Communications Technology Promotion funded by the Korea Government (MSIT) (Research on Adaptive Machine Learning Technology Development for Intelligent Autonomous Digital Companion) under Grant R-20161130-004520. This paper was recommended by Associate Editor Q. Meng. *(Corresponding author: Jong-Hwan Kim.)*

The authors are with the School of Electrical Engineering, Korea Advanced Institute of Science and Technology, Daejeon 34141, South Korea (e-mail: uhkim@rit.kaist.ac.kr; johkim@rit.kaist.ac.kr).

Color versions of one or more of the figures in this paper are available online at http://ieeexplore.ieee.org.

Digital Object Identifier 10.1109/TCYB.2018.2882921

spaces of individuals, such as home, have become connected and digitized. Smart Home refers to a digitized home that understands the inhabitants. Smart Home aims to provide personalized services to the residents as a housemaid. Smart Home reduces the burden of domestic labor of family members and maximizes relaxation and comfort in the home, thereby augmenting the functions of the house. Smart Home is emerging as a next-generation's core technology that improves human welfare and is drawing explosive responses in the market [6].

Two main technologies are leading the development of Smart Home: 1) service robots and 2) Internet of Things (IoT). A service robot performs useful tasks on behalf of humans with a semi or full autonomy. The tasks performed by a service robot do not include industrial automation applications in general [7]. Service robots applied in home environments support elderly people's activity of daily living [8], rehabilitation of the impaired [9], and daily household chores [10], [11]. In IoT frameworks, multiple sensor nodes offer the ability to measure the physical quantities of the environment, participate in the Internet network, and share the information regarding the observations [12]. Smart Home empowered by IoT allows users to control home appliances remotely and monitor the home environment in real time.

In this paper, we 1) design a stabilized memory system with a feedback mechanism for incremental learning and 2) propose an integrated robot-IoT framework for home service provision. For the design of a memory system, we base our design on an adaptive resonance theory (ART) network [30]–[32]. With the capability of adaptive pattern recognition and robust handling of temporal–spatial relations, the ART network can fully take the role of intelligence required to learn and infer a personalized service. However, the conventional ART network shows a few limitations: instability during the learning process and lack of feedback mechanism. We analyze the instability of the learning process mathematically and improve the learning stability by proposing an adaptive memory decay factor. Furthermore, the need for reflecting user feedback arises when the memory architecture reasons services for a user. To take the user feedback into account, we design a feedback mechanism which modulates memory parameters of a memory architecture so that memories of user preferred services get strengthened and memories of unsatisfactory services get weakened. We name the proposed memory architecture a stabilized feedback (SF) ART network.

In the proposed service framework, robot and IoT systems function as one unit to achieve a certain goal, and the proposed SF-ART takes the role of a learning and reasoning module. The work flow of the proposed framework is as follows. Robot and IoT systems collect information together and pass the data to the data interpreter module. The data interpreter module processes the received data and recognizes user actions and environment states. From this, the memory unit learns the pattern of user behaviors and the environment. After the learning procedure, the memory unit determines an appropriate service for the given user and environment states. Finally, robot and IoT systems collaborate to provide the service.

The main contributions of this paper are as follows.

1) We analyze the instability of the conventional memory systems. We mathematically formulate the learning procedure and show the limited performance of the conventional memory systems.

2) We propose an adaptive decay factor to enhance the stability of the learning process of the memory architecture and show the stabilized learning process of the proposed memory architecture.

3) We design a feedback mechanism to reflect user feedbacks. The feedback mechanism modulates the memory parameters so that the memory architecture develops in the manner a user prefers.

4) We propose a home service provision framework for robot and IoT collaboration using the proposed SF-ART architecture. In the framework, the memory architecture learns and reasons personalized services, and the two systems work as one unit to offer appropriate services.

5) We set up a Smart Home environment and verify the effectiveness of both the proposed memory architecture and the service framework. For the verification, we conduct thorough experiments under various conditions with scenarios.

The remainder of this paper is organized as follows. In Section II, we review related works with the limitations of previous research studies. Section III introduces preliminary research backgrounds for better understanding. Section IV analyzes the limitations of conventional memory structures and proposes enhanced learning procedures with the design of feedback mechanism. Section V describes the proposed service framework and the accompanying algorithms. Sections VI and VII illustrate the experiment settings and the analysis of the results. Discussion and concluding remarks follow in Section VIII.

## II. RELATED WORK

A variety of research studies have contributed to the ongoing development of Smart Home. In this section, we review the major categories of the research related to home service provision with both the robot and IoT systems: 1) learning and reasoning algorithms for service provision and 2) the integration of robotics and IoT systems. In addition, we describe ART networks on which the proposed memory system is based.

### A. Learning and Reasoning Algorithms

The learning and reasoning algorithms take a crucial role in service provision since they learn user's preference and determine which service to offer in a given situation. A typical learning and reasoning process proceeds as follows. The Smart Home system observes the user and environment and delivers information regarding the user behavior and environment state to the learning module. The learning module interprets the information and extracts patterns to learn user behaviors that change over time and situation. After learning, the reasoning module determines which service to provide by observing the user behaviors and the environment up to the decision making point. The learning and reasoning algorithms reported in the literature include the rule-based system [13], hidden Markov model (HMM) [14]–[16], computational intelligence (CI) [17]–[20], and reinforcement learning (RL) [21]–[23]. Rule-based algorithms encode expert knowledge and use the engineered knowledge to reason services for given situations. They require heaps of manual work for knowledge engineering and a substantial amount of time depending on the domain complexity. HMM and CI algorithms statistically model the user behaviors and let the system designers to control the degree of model complexity. However, HMM and CI algorithms learn patterns in offline manners, so they cannot deal with dynamically varying user behaviors and environment states [24]. On the other hand, RL takes an online learning approach and sets the internal parameters incrementally. RL, however, suffers from the curse of dimensionality, thus requiring huge amounts of data and time for training [23].

### B. Integration Systems

The research on the integration of the robot and IoT systems arises due to the inherent limitations of both systems. Service robots equipped with external sensors lack sensing coverage as well as sensing capability [25]. Attaching more sensors or high performance sensors on a robot does not alleviate the problem since it increases the amount of computation and the payload of the robot [26]. In addition, IoT systems whose components are installed on a fixed position have limited mobility and manipulation capability. Thus, the connectivity of an IoT system and the services the system can provide are constrained. The research on the integration aims to complement each system by hiding the limitations with the other system. Aiding a robotic system with IoT enhances data collection capability [25], [26], while an IoT system assisted by a robot overcomes the limited mobility and manipulation [27], [28]. However, these integrated systems do not successfully elicit the potential synergies from the integration. They merely result in enhanced robot or IoT systems where one part supports the other without collaboration between the two [29].

### C. ART Networks

An ART network stores a series of consecutive events and the temporal–spatial relations among the events in the form of an episodic memory which belongs to a declarative memory. With incremental feature extraction and fast online clustering, the ART network can effectively learn and store new experiences as episodes and retrieve existing episodes with partial cues. Conventional ART networks, however, exhibit two major drawbacks in Smart Home service systems: 1) learning instability and 2) deficiency in reflecting user feedback. First of all, all of the memories in conventional ART networks can get removed during the learning process as the networks learn a number of episodes. The strengths of episodes stored in the episodic memory based on ART networks converge to the same value with an increased number of episodes, and the episodes become prone to deletion. In addition, conventional ART networks do not possess a feedback mechanism, which confines its use in reasoning a service for users. When the networks infer nonpreferred services, users just have to take them or a system manager has to handle the issue manually whenever the issue arises.

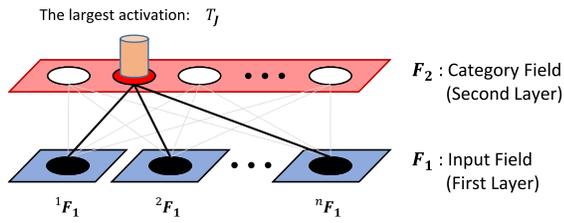

Fig. 1. Fusion ART structure. The structure consists of two layers: input field (first layer) and category field (second layer). The input field receives input vectors through multiple channels and conducts complement coding to avoid category proliferation.

## III. Preliminaries

We design the proposed memory architecture based on the Deep ART network which is built upon Fusion ART and EM-ART networks. For better understanding of our proposed memory architecture, we briefly introduce Fusion ART, EM-ART, and Deep ART structures.

### A. Fusion ART

Fig. 1 shows the Fusion ART structure. Fusion ART receives input vectors through multiple channels in contrast to Fuzzy ART which has a single input channel. Multiple input channels allow the network to group similar types of inputs into one category. The computational processes of Fusion ART are described in the following.

*1) Complement Coding:* Complement coding normalizes the input vector and prevents the category proliferation problem. Let $^k\mathbf{I} = (^kI_1, {}^kI_2, \ldots, {}^kI_d)$ denote the input vector for the $k$th channel $^kF_1$ of the input field and $d$ the dimension of the vector. Each input channel generates the corresponding *activity vector* $^k\mathbf{x} = (^k\mathbf{I}; {}^k\bar{\mathbf{I}})$, which concatenates the input vector $^k\mathbf{I}$ and its complement vector $^k\bar{\mathbf{I}} = \mathbf{1} - {}^k\mathbf{I}$.

*2) Code Activation:* The complement coded activity vectors go through the code activation process. The activity vector activates each category node in the $F_2$ layer by calculating the choice function

$$T_j = \sum_{k=1}^{n} {}^k\gamma \frac{|^k\mathbf{x} \wedge {}^k\mathbf{w}_j|}{\alpha + |^k\mathbf{w}_j|} \tag{1}$$

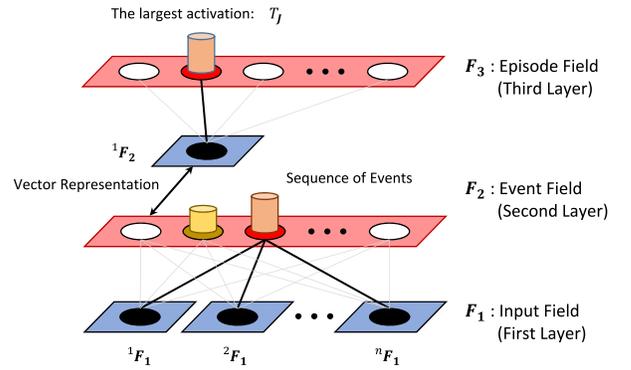

Fig. 2. Structure of EM-ART. The structure is built upon a Fusion ART structure. The episode field, the third layer, learns the temporal sequence of events. The $^1F_2$ represents the events occurred in $F_2$ as a vector.

where $^k\gamma \in [0, 1]$ is a contribution parameter that weighs each input channel's contribution to the choice function, $^k\mathbf{w}_j$ is a *weight vector* that connects the $j$th category node and $k$th input channel, $\alpha = 0^+$ is a choice parameter for numerical stability, $\wedge$ is the fuzzy AND operator defined as $(\mathbf{p} \wedge \mathbf{q})_i \equiv \min(p_i, q_i)$, and $|\cdot|$ is the norm defined as $|\mathbf{p}| \equiv \sum_i p_i$.

*3) Code Competition:* Code competition selects the category node that best represents the activity vector

$$J = \arg\max_{j \in S_{F_2}} T_j \tag{2}$$

where $S_{F_2} = \{1, 2, \ldots, n_C\}$ contains the indices of category nodes in the $F_2$ layer and $n_C$ is the number of category nodes. Here, the activation value is considered as the degree of similarity.

*4) Template Matching:* The chosen category node passes through the template matching process which tests if the category node and the input vector meet the resonance condition

$$^km_J = \frac{|^k\mathbf{x} \wedge {}^k\mathbf{w}_J|}{|^k\mathbf{x}|} \geq {}^k\rho \tag{3}$$

where $^k\rho$ is a vigilance parameter which works as the resonance threshold. If the resonance condition is not met, a new category node is created and the associated weight vector is initialized as $^k\mathbf{x}$.

*5) Template Learning:* If the node satisfies the resonance condition, learning occurs and the weight vector is updated as follows:

$$^k\mathbf{w}_J^{(\text{new})} = (1 - \beta)^k\mathbf{w}_J^{(\text{old})} + \beta\left(^k\mathbf{x} \wedge {}^k\mathbf{w}_J^{(\text{old})}\right) \tag{4}$$

where $\beta \in [0, 1]$ is a learning rate.

### B. EM-ART

EM-ART encodes a temporal sequence of events using two concatenated ART networks as depicted in Fig. 2. The second layer which is a category layer in the Fusion ART structure is an event layer in EM-ART. The event layer represents a temporal sequence with a decay factor, and the episode layer, the third layer, categorizes the sequence as an episode.

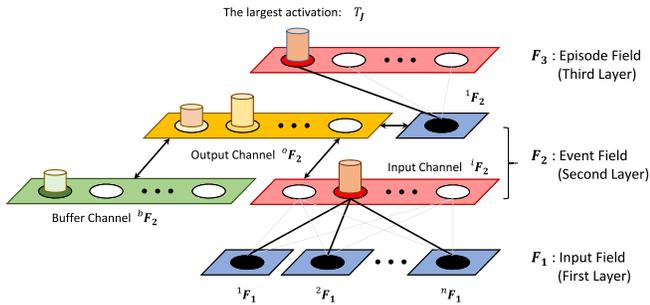

Fig. 3. Deep ART network. The structure uses additional buffer and output channels to differentiate the same events in a sequence and reduce the retrieval error. These additional channels do not increase the computation.

*1) Episode Encoding:* As a node in the event layer gets fired by an input vector, the event layer develops as follows:

$$y_j^{(\text{new})} = \begin{cases} 1, & j = J \\ y_j^{(\text{old})}(1 - \tau), & j \neq J \end{cases} \tag{5}$$

where $\tau \in (0, 1)$ is a decay factor. The output of the event layer becomes $\mathbf{y} = (y_1, y_2, \ldots, y_{n_C})$. Sorting the $y_i$'s in the descending order recovers the temporal sequence because every $y_i$ decreases at every update step except the current event indexed with $J$.

*2) Episode Learning:* The top ART network classifies and stores the temporal information encoded in the vector $\mathbf{y}$. The episode layer only consists of a single channel and accepts the vector $\mathbf{y}$ through the channel. The same matching and learning processes occur and the choice function is as follows:

$$T_p = \frac{|\mathbf{y} \wedge \mathbf{w}_p|}{|\mathbf{w}_p|} \tag{6}$$

where $\mathbf{w}_p$ is the weight vector for the $p$th episode node in the episode field $F_3$, the third layer.

*3) Memory Strength:* EM-ART employs the concept of memory strength for an efficient memory management. Frequently activated memories get strengthened with a reinforcement factor and infrequent memories get weakened

$$s_j^{(\text{new})} = \begin{cases} s_{\text{init}}, & \text{created} \\ s_j^{(\text{old})} + \left(1 - s_j^{(\text{old})}\right)r, & \text{reactivated} \\ s_j^{(\text{old})}(1 - \delta), & \text{otherwise} \end{cases} \tag{7}$$

where $s_{\text{init}}$ is an initial value of the memory strength, $r \in [0, 1]$ is a reinforcement factor, and $\delta \in [0, 1]$ is a decay factor. If $s_j$ falls below a threshold $\theta \in [0, 1]$, the $j$th node gets deleted from the memory.

### C. Deep ART

EM-ART causes episode retrieval errors when episodes have different lengths, or the same events are repeated in an episode [32]. Deep ART shown in Fig. 3 employs additional channels and enhanced time sequence encoding and decoding methods to overcome the limitation, without additional channels increasing the computational complexity.

*1) Episode Encoding:* The buffer channel buffers the output channel vector. In calculating the new output vector, the previous output vector stored in the buffer channel gets combined with the input. This process differentiates the same event repeated in an episode. Let $^{\mathbf{i}}\mathbf{y}$, $^{\mathbf{b}}\mathbf{y}$, and $^{\mathbf{o}}\mathbf{y}$ denote the input, buffer, and output vectors, respectively. The input vector $^{\mathbf{i}}\mathbf{y}$ represents the current activated node in the event layer as follows:

$$^i y_j = \begin{cases} 1 & j = J \\ 0 & j \neq J. \end{cases} \tag{8}$$

The buffer channel stores the previous output vector $\mathbf{y_o}$, i.e., $\mathbf{y_b} = {}^{\mathbf{o}}\mathbf{y}^{(\text{prev})}$. The output vector $^{\mathbf{o}}\mathbf{y}$ combines the input vector and the buffer vector as follows:

$$\begin{aligned} {}^{\mathbf{o}}\mathbf{y}^{(\text{new})} &= {}^i w \cdot {}^{\mathbf{i}}\mathbf{y} + {}^b w \cdot {}^{\mathbf{b}}\mathbf{y} \\ &= {}^i w \cdot {}^{\mathbf{i}}\mathbf{y} + {}^b w \cdot {}^{\mathbf{o}}\mathbf{y}^{(\text{old})} \end{aligned} \tag{9}$$

where $^i w$ and $^b w$ are weights of the input channel and buffer channel, respectively.

*2) Episode Retrieval:* The retrieval of an episode from Deep ART takes six steps: 1) select the episode index $J$; 2) readout the corresponding weight vector $w_j$ of the $j$th episode; 3) extract the index of the largest element in $^{\mathbf{o}}\mathbf{y}$, i.e., $^{\mathbf{o}}\mathbf{y}_{\max}$; 4) store the index in a sequence queue; 5) subtract the largest $w_i \cdot (w_b)^n < {}^o\mathbf{y}_{\max}$ from $^o\mathbf{y}_{\max}$ for a positive integer $n$; and 6) repeat steps 3)–5) until no positive element remains in $^{\mathbf{o}}\mathbf{y}$. Then the indices in the sequence queue represents the temporal sequence of events in an episode.

## IV. Stabilized Feedback ART

Although the conventional ART networks are effective unsupervised and incremental clustering algorithms, they go through instable learning stages for specific learning settings. Furthermore, the conventional ART networks do not support the feedback mechanism; thus, they cannot reflect user preferences effectively when used to reason proper user services. In the following, we first analyze the learning instabilities using a mathematical proof. Then, we propose a technique to enhance the learning stability of the memory architecture. Next, we design a feedback mechanism so that the system can develop based on the user preference.

### A. Enhancement of Stability

The original paper of EM-ART [31] and Deep ART [32] does not mention the guidelines for setting a decay factor for memory strength and the threshold for removing a memory element. Those who use EM-ART or Deep ART have to nudge the values for multiple times to find the proper values. Furthermore, the decay factor is fixed throughout the whole process of EM-ART and Deep ART. We analyze the potential risk of setting a fixed value for the decay factor and provide set-up guidelines for the decay factor and the threshold.

To find proper values for the decay factor and threshold, we start by analyzing the equilibrium state. In the equilibrium state, the memory strength of each episode retains its value over iterations as the memory system achieves stability.

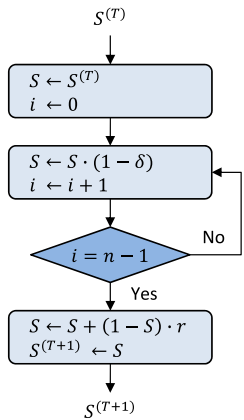

Fig. 4. Computation flowchart of memory strength for a single iteration. The memory strength decays $(n-1)$ times and gets reinforced once.

Let $s$, $\delta$, $r$, and $n$ denote the memory strength, decay factor, reinforcement factor, and the number of episodes in the memory network, respectively. $n$ is a system variable of the memory system which can be retrieved from the system, and the variable dynamically changes as the memory system learns or forgets episodes. To achieve the equilibrium state, every episode in the memory gets activated equally once per $n$ activations, and a single iteration consists of $n$ activations. Then, each episode in the network goes through the decay process $(n-1)$ times and the reinforcement process at the $n$th step in every iteration as in Fig. 4. By indicating the iteration number with superscription T, the memory strength after one iteration becomes

$$s^{(T+1)} = s^{(T)}(1-\delta)^{n-1} + \left(1 - s^{(T)}(1-\delta)^{n-1}\right)r \quad (10)$$

which can be rewritten as a recurrence relation

$$s^{(T+1)} = r + (1-r)(1-\delta)^{n-1}s^{(T)}. \quad (11)$$

To solve the recurrence relation, we transform (11) as follows:

$$s^{(T+1)} - \frac{r}{1-\Delta} = \left(s^{(T)} - \frac{r}{1-\Delta}\right) \cdot \Delta \quad (12)$$

where $\Delta = (1-r)(1-\delta)^{n-1}$. Then, the memory strength at the $T$th iteration turns into

$$s^{(T)} = \left(s_{\text{init}} - \frac{r}{1-\Delta}\right) \cdot \Delta^{(T-1)} + \frac{r}{1-\Delta}. \quad (13)$$

We consider two terminal points. First, $s^{(T)} \rightarrow r$ as $n \rightarrow \infty$ and $T \rightarrow \infty$. From this, we can induce that the threshold value for memory deletion should be set larger than the value of the reinforcement factor for meaningful operations. Otherwise, no element gets removed from the network. Second, $s^{(T)} \rightarrow [r/(1-\Delta)]$ when $T \rightarrow \infty$ and $n$ is fixed. The terminal value, $[r/(1-\Delta)]$, is relatively larger than $r$ so that the threshold value should be set accordingly for proper functioning. We investigate the numerical values in Section V.

For the alleviation of the instability, we propose an adaptive decay factor. We derive the adaptive decay factor by forcing a memory strength to return to the original value after a single iteration as follows:

$$s_{\text{init}}(1-\delta)^{n-1} + \left(1 - s_{\text{init}}(1-\delta)^{n-1}\right)r = s_{\text{init}}. \quad (14)$$

Factoring (14) produces

$$\frac{s_{\text{init}} - r}{s_{\text{init}}(1-r)} = (1-\delta)^{n-1}. \quad (15)$$

For (15) to be accountable for even large $n$, the number of episodes in the memory network, we take the limit on the right side

$$\frac{s_{\text{init}} - r}{s_{\text{init}}(1-r)} = \lim_{n \rightarrow \infty}(1-\delta)^{n-1}. \quad (16)$$

Setting $\delta = [\delta_{\text{init}}/(n-1)]$ for $n > 1$ in (16) yields

$$\frac{s_{\text{init}} - r}{s_{\text{init}}(1-r)} = \lim_{n \rightarrow \infty}\left(1 - \frac{\delta_{\text{init}}}{n-1}\right)^{n-1}$$
$$= \left(\frac{1}{e}\right)^{\delta_{\text{init}}} \quad (17)$$

which results in

$$\delta_{\text{init}} = \ln\left(\frac{s_{\text{init}}(1-r)}{s_{\text{init}} - r}\right). \quad (18)$$

Therefore, the adaptive decay factor $\delta = [\delta_{\text{init}}/(n-1)]$ with the initial value in (18) remedies the memory strength instability problem. In addition, an episode does not get forgotten if it is activated at least once per $n$, the number of episodes in the memory system, activations with the proposed adaptive decay factor.

In summary, the conventional memory strength update causes all memory strengths to converge to the same value in spite of regular activations when the network learns a number of episodes. The threshold for memory deletion should be set higher than the reinforcement factor for the effectiveness. Moreover, employing an adaptive decay factor which varies according to the number of episodes stored in the memory system with the initial value defined in (18) mitigates the convergence problem.

### B. Design of Feedback Mechanism

The proposed memory architecture learns user behaviors and infers a proper service based on the current user state and the environment after learning. Therefore, incorporating user's feedback into the provided services is necessary. Through feedback, appropriate services are strengthened, and improperly learned memories get attenuated. As a result, the user can gradually receive the desired services, and the user satisfaction increases over time.

TABLE I
TYPES OF FEEDBACK

| Feedback | Type | Content | Feedback Value | Effect on | |
|---|---|---|---|---|---|
| | | | | Memory Strength | Vigilance |
| Strong Positive | Explicit | Satisfaction | $> \xi_w$ | Reinforcement | Decay |
| Weak Positive | Implicit | Satisfaction | $= \xi_w$ | Reinforcement | None |
| Negative | Explicit | Dissatisfaction | $< \xi_w$ | Decay | Reinforcement |

TABLE II
MEMORY PARAMETER MODULATION FUNCTIONS

$$f_{memory}(s_j{}^{(old)}, \; \xi_j) = \begin{cases} min(1, \; s_j{}^{(old)} + 2(1 - s_j{}^{(old)}) \cdot r_s), & \xi_j > \xi_w \\ s_j{}^{(old)} \cdot (1 - \delta_s)^2, & s_j{}^{(old)} - s_{init} \cdot (1 - \delta_s)^p > 0 \\ \theta \cdot (1 - \delta_s)^{-1} & \text{otherwise} \end{cases} \quad (19)$$

$$f_{vigilance}(\rho_j{}^{(old)}, \; \xi_j) = \begin{cases} \rho_j{}^{(old)} \cdot (1 - \delta_\rho), & \xi_j > \xi_w \\ min(1, \; \rho_j{}^{(old)} + (1 - \rho_j{}^{(old)}) \cdot r_\rho), & \text{otherwise} \end{cases} \quad (21)$$

*1) Types of Feedback:* In order to handle the user feedback more efficiently, the proposed memory architecture accepts three types of feedback from the user. Table I summarizes the types of feedback.

1) *Strong Positive Feedback:* The strong positive feedback refers to the explicit expression of satisfaction from the user. This expression works as a compliment and strengthens the memory of the provided service.
2) *Weak Positive Feedback:* For the weak positive feedback, the user does not explicitly express any opinion regarding the offered service, but just accepts the service. The acceptance of a service is considered as an implicit expression of satisfaction.
3) *Negative Feedback:* The user gives the negative feedback when the user is not satisfied with the service, or the service is not what the user intended. From the negative feedback, the memory component for the service gets weakened.

*2) Memory Parameter Modulation:* We propose to modulate the memory strength and vigilance parameter on receiving feedback from users. The memory strength determines how long a memory component stays within the memory network, and the vigilance parameter controls the likelihood of activation or retrieval of a memory component.

*a) Memory strength modulation:* The strong positive feedback and negative user feedback modulate the strength of the memory component associated with the service user has received, as in (19) in Table II. The strong positive feedback enhances the memory strength with a doubled reinforcement factor. The negative feedback attenuates the memory strength in two manners. First, the negative feedback reduces the memory strength twice if the memory has been activated at least once in the last $p$ episodes (we name the parameter, $p$, episode parameter). In the second case, the negative feedback makes the memory strength become $\theta \cdot (1 - \delta_s)^{-1}$, and the memory will get deleted if it is not activated in the next retrieval. The weak positive feedback works as activation for memory strengths. The memory strength decays normally with

no feedback given. Finally, the memory strength develops as

$$s_j{}^{(new)} = \begin{cases} s_{init}, & \text{created} \\ s_j{}^{(old)} + (1 - s_j{}^{(old)})r, & \xi_j = \xi_w \\ f_{memory}(s_j{}^{(old)}, \; \xi_j), & \xi_j \neq \xi_w \\ s_j{}^{(old)}(1 - \delta_s), & \text{otherwise} \end{cases} \quad (20)$$

where $\xi_j$ is a feedback received from a user, $j$ is the index of the episode for the service, and $\xi_w$ is the value of the weak positive feedback set by a system designer.

*b) Vigilance parameter modulation:* The strong positive feedback lowers the threshold for the activation and the negative feedback does the opposite, as in (21) in Table II.

The development of the vigilance parameter over the time becomes

$$\rho_j{}^{(new)} = \begin{cases} \rho_{init}, & \text{created} \\ f_{vigilance}(\rho_j{}^{(old)}, \; \xi_j), & \xi_j \neq \xi_w \\ \rho_j{}^{(old)}, & \text{otherwise.} \end{cases} \quad (22)$$

Although the vigilance parameter allows the memory component to pass the template matching process with a lower threshold value, the memory component will not reach the template matching process if it does not win the code competition over similar memory components. We propose the following choice function for code activation to let the memory components that have received strong positive/negative feedback from the user have competitive advantage/disadvantage

$$T_j = \left( \frac{\rho_{init}}{\rho_j} \right) \cdot \sum_{k=1}^{n} {}^k\gamma \cdot \left( \frac{|{}^k\mathbf{x} \wedge {}^k\mathbf{w}_j|}{\alpha + |{}^k\mathbf{w}_j|} \right). \quad (23)$$

The coefficient $(\rho_{init}/\rho_j)$ accounts for the competitive advantage/disadvantage for a memory component that received the strong positive/negative feedback.

## V. COLLABORATIVE ROBOT-IoT FRAMEWORK

In this section, we propose a framework for service provision by robot and IoT collaboration using the memory architecture designed in the previous section as a learning and reasoning module. Fig. 5 shows the framework architecture diagram.

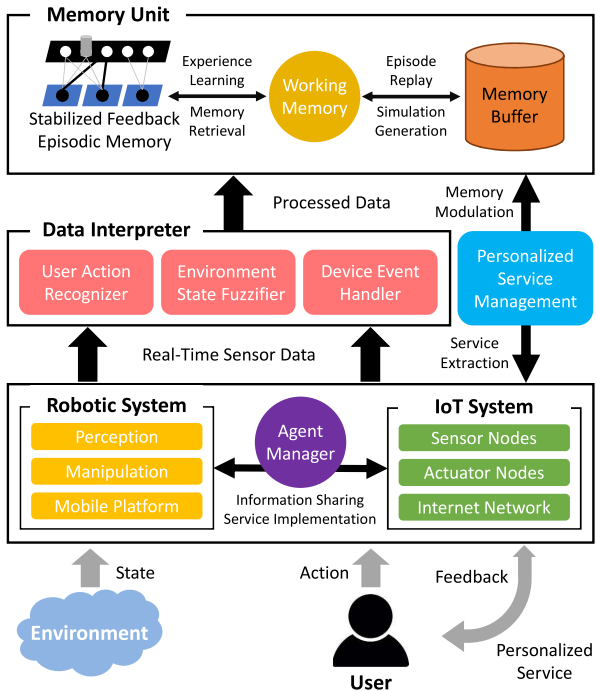

Fig. 5. Proposed home service provision framework. In the framework, the robot and IoT systems work as one unit to gather information and to provide home services. The proposed SF-ART plays the role of learning and reasoning.

### A. Robot and IoT Systems

In the proposed framework, the robot and IoT systems function as a single integrated unit. One system completes tasks which the other system is incompetent in, counterbalancing the overall workload.

1) The perception part in the robot and the sensor node in the IoT system gather information and detect changes in the user state and the surrounding environment collaboratively. The robot conducts active perception, such as vision with tracking and simultaneous localization and mapping, and the IoT system collects information regarding device usage and environment states.

2) The manipulation part in the robot and the actuator node in the IoT system exert a physical effect on the environment or provide a personalized service. The robot performs tasks which require mobility, such as moving an object, disposing garbage, making cereal, and doing home chores. On the other hand, the IoT system directly operates on the device, such as lifting blinds, turning on/off stand-light, and cooking in a microwave. It is difficult for the IoT system to perform tasks requiring mobility which are usually executed by the robot since IoT is fixed at one point, while the operations the IoT system conducts would demand complex task planning and precise control for the robotic system.

3) The mobile platform of the robotic system enables the robot: a) to actively sense the user and the environment and b) to provide services which require movement and displacement to implement. Active perception helps the agent to effectively collect relevant and crucial information for service provision. A manipulator equipped with a mobile platform can offer services with more versatility.

4) The Internet network of the IoT System allows communication with external devices and the user. It can not only send messages containing the information on the internal states but also receive commands or information from external systems. By expanding the network coverage in the framework, the whole framework can also communicate with external systems.

### B. Data Interpreter

The data interpreter receives raw data from robot and sensor networks and processes the raw data to extract meaning. To provide the best appropriate service to the user, the interpreter extracts three kinds of information. The following lists the types of information the unit extracts.

1) *User Action:* The service to be provided for the user depends on the action taken by the user. For example, Smart Home can provide a service that turns off the restroom light when the user leaves the restroom. In other words, the user intention can be inferred through the action taken by the user. Thus, the user action recognizer recognizes the user's actions to infer the intention of the user. Examples of user behaviors include returning home, going out, entering the bathroom, and using a microwave oven. Sensors collecting information on location and movement can aid the user action recognition process.

2) *Environment State:* Smart Home collects the environment information because the services the user desires vary depending on the environmental conditions. For example, a user wishes an air conditioner to lower the temperature when it is hot and to raise the temperature when it is cold. Environmental information collected by the proposed framework includes temperature, humidity, illumination, and noise level. The framework expresses the environmental information using fuzzy variables (environment state fuzzifier). Because environmental conditions felt by humans, such as hotness and coldness, differ from person to person, fuzzification can handle the vagueness efficiently.

3) *Device Information:* A device event handler extracts information regarding the changes in states of devices that directly provide a service to the user. Understanding how the user uses a device in a specific situation can allow the device to autonomously serve the user when the user is later in the same or similar situation. For example, if Smart Home learns the situation where the user enters the house when it is dark and turns on the light, Smart Home can provide a service to turn on the light when the user enters the house when it is dark. The main jobs of the handler are twofold: a) device event encoding and b) device state encoding. The device event represents an action exerted on a device, such as turning on/off a light, and the device state represents the internal state, such as on/off state of a light.

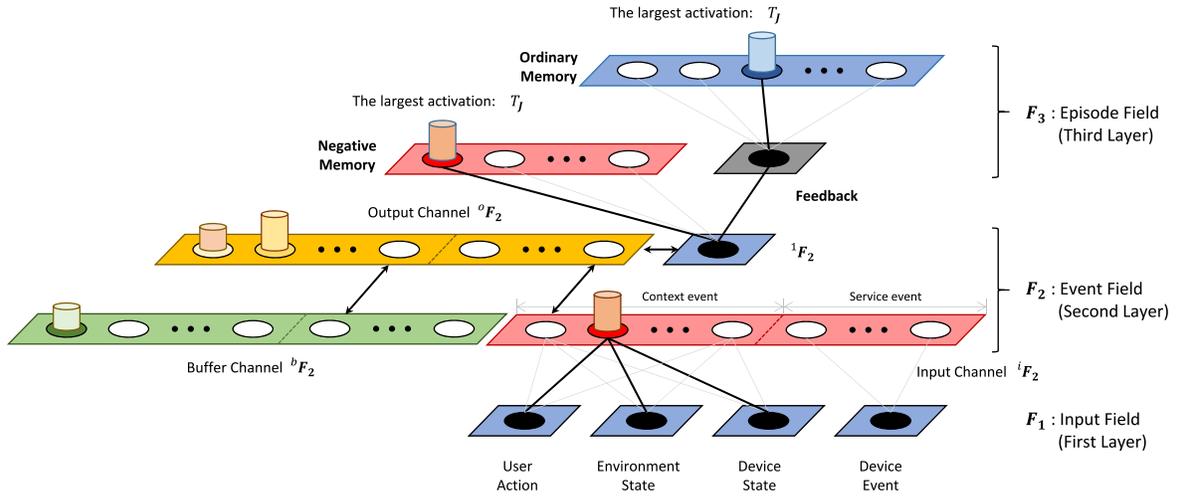

Fig. 6.   SF-EM network. The basic building block of SF-EM is SF-ART. The top layer stores two types of memories: ordinary and negative memory. By storing negative memory, SF-EM can fix inappropriately learned memories and extract user desired services.

### C. Memory Unit

The memory unit receives information from the data interpreter and learns the user behavior according to the environment. After learning, the memory unit infers the service to provide for the user based on the user behavior observed up to the decision making point. SF episodic memory (SF-EM) is designed using the proposed SF-ART memory architecture as the basic building block and the sequence encoding scheme of Deep ART. Fig. 6 shows the architecture of the SF-EM. The top layers of SF-EM consist of two parts: 1) the ordinary memory and 2) the negative memory. The ordinary memory stores ordinary episodes, while the negative memory stores episodes that received negative feedback from the user. The negative memory prevents the memory unit from repeatedly learning and retrieving episodes that received negative feedback. The event layer of SF-EM and the negative memory do not employ the feedback mechanism and the memory strength development to let the memory learn as various types of inputs as possible. On the other hand, the ordinary memory employs both the feedback mechanism and the memory strength development to reflect user preferences.

The SF-EM receives inputs through four channels each related to the user, environment, device state, and device event, respectively. We treat user action, environment state, and device state as one unit (context) and the device event as the other unit (service). This is to manage the information effectively. For example, the framework can provide the user with the services, such as "turning on the light" right after user comes home when it is dark and the light is off. The device event indicates the service, turning on the light, and the user, environment state, and device state together represent the context. The two units are independent and one unit is a zero vector when the other is not. In other words, the input has the form of either $\mathbf{x} = [\mathbf{c}; \mathbf{0}]$ or $\mathbf{x} = [\mathbf{0}; \mathbf{s}]$, where $\mathbf{c}$ and $\mathbf{s}$ stand for the context vector and service vector, respectively.

Another major point of training SF-EM is the ability to detect at which event a new episode starts and at which event the episode ends. The working memory in the memory unit (Fig. 5) recognizes the start and the end of an episode through temporal information. Algorithm 1 depicts the complete flow of the episode recognition. The episode recognition algorithm combines two events if they occur sequentially. The algorithm utilizes memory buffer which contains previous events. After inspecting the events in the memory buffer, which we name simulation generation, the algorithm merges two events that occur sequentially. If a pair of events occur just once in the buffer, the algorithm does not associate the two and no repetition of subsequence is allowed within an episode. Furthermore, the algorithm inserts a time gap event between two events if

---

**Algorithm 1** Episode Recognition

**Input**: Memory Buffer, memBuff
**Output**: List of Episodes

1: event_queue = []
2: episode_list = []
3: **for** $i$ = 1 to memBuff.*length* **do**
4:     event_queue.append(classify(memBuff[$i$]));
5: **end for**
6: episode_list = event_queue;
7: **while** True **do**
8:     // update episode_list
9:     **for** each pair of episodes in episode_list **do**
10:         **if** occurred sequentially $\geq 2$ **then**
11:             merge and insert time gap
12:         **end if**
13:     **end for**
14:     **if** no update **then**
15:         break;
16:     **end if**
17: **end while**

---

**Algorithm 2** PSM

**Given**: Service Episode $\mathbf{y} = (e_1, e_2, ..., e_k)$
**Effect**: Service Provision

```
 1: for i = 1 to y.length − 1 do
 2:    if y[i].isComplete() then
 3:        continue;
 4:    end if
 5:    switch y[i].type() do
 6:        case "user":
 7:            wait(y[i]);
 8:        case "device":
 9:            implement(y[i]); // provide service
10:            if feedback ≠ Null then
11:                modulate(y[i], feedback);
12:                if feedback < ξ_w then
13:                    // Stop servicing and learn new episode
14:                    readyForNegativeFB(y[0 : i − 1]);
15:                    break;
16:                end if
17:            end if
18: end for
```

the two events appear with a regular time gap. The algorithm connects the chunks of events in the same way, which refers to the process of episode replay.

### D. Personal Service Management

Personal service management (PSM) controls the service provision procedure. Algorithm 2 shows the complete flow of PSM. After the memory unit retrieves the personalized service sequence and delivers the service routine to PSM, PSM implements each service at the appropriate time. For example, assume that the user is lying on the bed and that the memory unit has extracted the following episode: the user 1) lies on the bed, 2) turns off the lights and lowers the blinds, 3) sleeps, 4) raises the blinds, 5) prepares the cereal, 6) has breakfast, and 7) cleans up. PSM will implement the second event to provide the service when the first event occurs. After that, PSM waits for the end of the third event and implements the events 4 and 5 to provide services. Then, PSM waits for the sixth event to finish and delivers the last service. If the user gives feedback in the middle, PSM passes the feedback to the memory unit so that the memory parameter modulation starts operating. For the negative feedback, the service routine stops, and the rest of user behaviors are observed to learn a new episode. The current episode is stored in the negative memory of SF-EM. This corrects the inappropriately learned episodes.

## VI. SIMULATION

In this section, we verify the performance of SF-ART's two features: 1) adaptive decay factor and 2) feedback mechanism. These features enhance the learning stability and the functionality of the conventional ART. We conduct a series of simulations and compare SF-ART against conventional ART to demonstrate the superior performance of SF-ART.

### A. Adaptive Decay Factor

For the validation of the adaptive decay factor, we conducted a pair of simulations. In the first simulation, we trained SF-EM based on SF-ART and Deep ART with certain numbers of episodes and then observed the development of memory strength over iterations. A single iteration consists of fixed $n$ activations, where $n$ denotes the total number of episodes in a network. Each episode gets activated once per $n$ activations, i.e., once a single iteration and decayed $(n - 1)$ times in an iteration. We varied the following network parameters: the number of episodes (10, 20, 50, and 100), the value of decay factor (0.01 and 0.05), and the value of reinforcement factor (0.05 and 0.1). Other parameters, such as the vigilance parameter and similarity measure were kept the same for fair comparison. The learning parameters were as follows: value of vigilance parameter ($\rho$) = 0.9, contribution parameter ($\gamma$) = 0.25, learning rate ($\beta$) = 0.5, value of choice parameter ($\alpha$) = 0.01, initial memory strength ($s_{init}$) = 0.8, dimension of input vector = 2, and number of channel = 1. Other parameters, though set equal to those used in the following section, did not affect the result of the first simulation in this section.

Fig. 7 shows the simulation results of a single memory strength, where the horizontal axis and the vertical axis indicate the number of activations and the memory strength, respectively. The top two lines described by only $r$ in the legend represent the memory strength development with the proposed adaptive decay factor. The value of the proposed adaptive decay factor depends on the number of memories in the network ($\delta_{init}/[n - 1]$), and the initial value is determined by the initial memory strength and the reinforcement factor, as in (18). The proposed adaptive decay factor kept the memory strength stable as the memory component was activated regularly. The memory strength decayed gradually and recovered at the end of each iteration. The adaptive decay factor properly worked for all simulation settings. As the memory parameter changed, the proposed decay factor adaptively varied its value to cope with the different parameters. However, the conventional memory development led all the memory components to decrease and converged to the value, $[r/(1 - \Delta)]$ in spite of regular activations. As the number of episodes in the network increased, the memory strengths got attenuated sharply and were not able to recover. Thus, the conventional memory development resulted in the memory strength instability, i.e., deletion of all memory components.

The terminal values vary according to the number of categories in the network, reinforcement factor, and decay factor. We repeat the following equation for clarity:

$$\text{terminal value} = \frac{r}{1 - \Delta} = \frac{r}{1 - (1 - r)(1 - \delta)^{n-1}}. \quad (24)$$

We computed the terminal values for the simulation environments in Table III. For small $n$'s, the terminal values are relatively large because the time gap between two activations is shorter for smaller $n$'s. Accordingly, the appropriate threshold value for forgetting should be set higher than the terminal value. In this situation, the memory network becomes too sensitive and deletes memories frequently. By contrast, the terminal values for large $n$'s converge to the reinforcement factor $r$ which is much smaller than the terminal values of smaller $n$'s. The time gap between two activations is longer for large $n$'s, and the memory strengths do not recover. In this situation, every memory component gets deleted.

TABLE III
TERMINAL VALUES

| $(d, r)$ | $n$ | | | |
|---|---|---|---|---|
| | 10 | 20 | 50 | 100 |
| $(.01, .10)$ | 0.5623 | 0.3899 | 0.2222 | 0.1499 |
| $(.01, .05)$ | 0.3783 | 0.2324 | 0.1192 | 0.0771 |
| $(.05, .10)$ | 0.2317 | 0.1514 | 0.1079 | 0.1006 |
| $(.05, .05)$ | 0.1246 | 0.0779 | 0.0547 | 0.0503 |

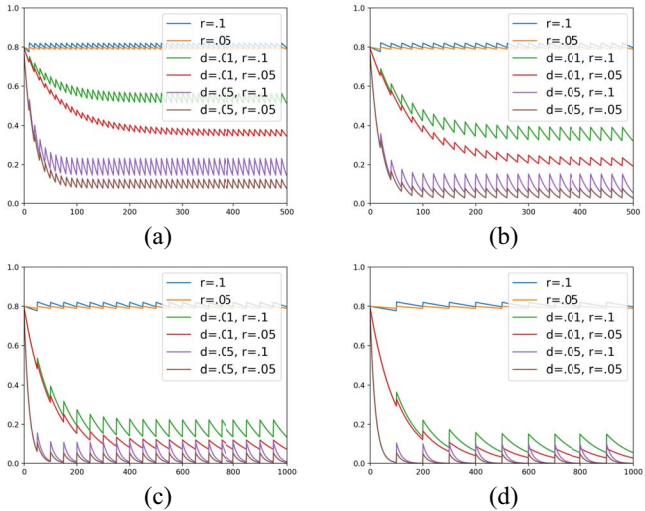

Fig. 7. Simulation result of the memory strength development (fixed $n$). The memory strength develops stably with the proposed adaptive decay factor. On the other hand, the conventional memory strength development leads to low terminal values despite regular activations. (a) $n = 10$. (b) $n = 20$. (c) $n = 50$. (d) $n = 100$.

In the second simulation for the adaptive decay factor, we dynamically varied the number of episodes in the memory. To verify the performance of the adaptive decay factor in various environments, we also changed the initial value of the memory strength and the value of the reinforcement factor. Other parameters were kept intact as in the first simulation. Fig. 8 shows the simulation result with the dynamic variation of the number of episodes, $n$, in the network. The memory strengths stay within a boundary, and they do not converge or diverge in all simulation environments. It is summarized that with a higher value of reinforcement factor, the fluctuation of boundary becomes larger. The second simulation also verifies the proposed adaptive decay factor which stabilizes the memory system.

### B. Feedback Mechanism

To verify the effectiveness of the proposed feedback mechanism, we compared the memory networks equipped with the feedback mechanism against the conventional ART network. We considered three memory architectures: 1) SF-EM with both the feedback mechanism and negative memory; 2) Deep

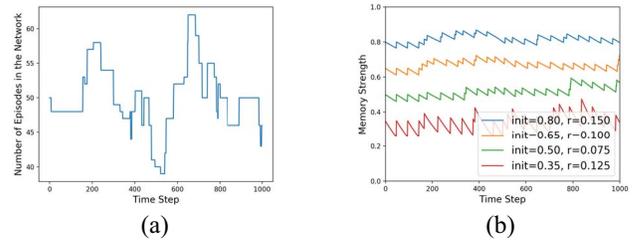

Fig. 8. Simulation result of the memory strength development for dynamic $n$. As $n$ dynamically varies, the adaptive decay factor adaptively handles the memory strengths.

ART with the feedback mechanism; and 3) vanilla Deep ART. Deep ART with the feedback mechanism did not possess negative memory. We trained each network with the same simulation scenario. The simulation scenario consists of two phases. In the first phase, we put a couple of events in the memory buffer in advance and let the network learn service episodes. Then, we fed a sequence of events into each memory architecture and compared the extracted services from each memory network. We used one event as a partial cue to retrieve an episode at each step. In the second phase of the simulation, we changed the user preference and traced how each memory network reacted to the variation. After the change of user preference, we kept feeding each network with the same partial cue to observe the development of the reaction each network showed over time. The learning parameters were as follows: the same as those used in the previous section except decay and reinforcement factors, value of strong positive feedback $= 2$, value of weak positive feedback $= 1$ ($= \xi_w$), value of negative feedback $= -1$, episode parameter ($p$) $= 2$, and value of reinforcement factor ($r$) $= 0.1$.

Events were denoted by alphabets, and the initial event queue in the memory buffer was "$a \rightarrow b \rightarrow c \rightarrow d \rightarrow a \rightarrow b \rightarrow c \rightarrow d \rightarrow e \rightarrow f \rightarrow g \rightarrow e \rightarrow f \rightarrow g$." From this event queue, two episodes were learned by the proposed episode recognition algorithm (Algorithm 1)

$$\text{episode } 1 : a \rightarrow b \rightarrow c \rightarrow d$$
$$\text{episode } 2 : e \rightarrow f \rightarrow g. \qquad (25)$$

Fig. 9 shows the simulation process and the result after the initial training. The order of event is marked above each event circle in the diagram. When the first event $a$ was fed into each network as a partial cue, episode 1 was retrieved from all of the memory networks. The PSM provides the rest of episode 1 as a service. Events $b$ through $d$ stand for the service routine for episode 1. Then, event $e$ was fed in at the fifth place, and each memory unit extracted episode 2. The corresponding service routine was offered by the PSM (event $f$ through event $g$). After the service routine for episode 2, we simulated a change in user preference to consider a case where a user starts to prefer a different service routine. On observing event $e$ at the eighth place, the memory architectures extracted episode 2 again, but the user rejected the service. For the first two memory architectures which have a feedback mechanism, user gave negative feedback and trained the memory architectures with a new episode by showing the newly preferred service routine. For the Deep ART network which does not have a feedback mechanism, the user stopped the service and simply showed the new service routine. The memory architectures learned the new episode: event $e$ plus event number 10 through 12 (episode 3). After the 12th event, the three memory architectures exhibited different behaviors.

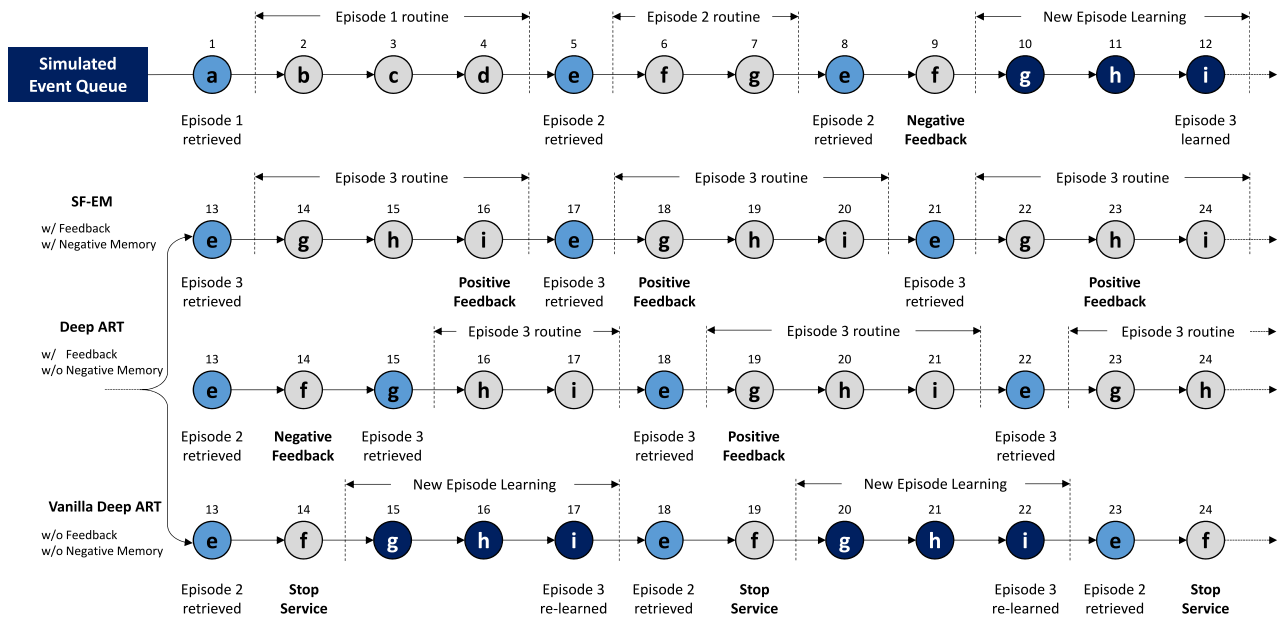

Fig. 9. Flowchart of the simulation process of the feedback mechanism and the result after the initial training. The order of event is marked above the circle representing each event. After receiving negative feedback from the user, the proposed feedback mechanism and the negative memory let the memory network react coherently, while the conventional memory network fails to reflect the change in user preference.

First, the proposed SF-EM which has both the feedback mechanism and negative memory reacted instantly to the negative feedback and offered episode 3 rather than episode 2 as a service upon observing event $e$ at the 13th place. This can be explained in the following. After learning both episode 2 and episode 3, the memory architecture retrieved episode 2 with the partial cue of event $e$ since episode 2 shorter than episode 3 results in bigger code activation. However, SF-EM had stored episode 2 in the negative memory at the ninth place and the negative memory blocked the system to retrieve episode 2 as a service. The next strong memory, episode 3, was instead retrieved. Then, the user received the desired service. The user could provide positive feedback at any place during the service routine.

Next, Deep ART with the feedback mechanism which does not have negative memory reflected the changed user preference after serving episode 2 once. As the episode parameter ($p$) was set to 2, episode 2 got deleted after receiving second negative feedback. Before episode 2 was removed from the memory architecture, it kept outputting episode 2. After episode 2 was removed from the network, Deep ART extracted the correct episode, episode 3.

Finally, vanilla Deep ART which does not have neither feedback mechanism nor negative memory could not manage the change in user preference. It kept on providing episode 2 as a service upon observing the partial cue of event $e$. Vanilla Deep ART just had to wait for episode 2 to get removed from the memory network. In addition, the user could not feed negative feedback into the system and just had to stop the system when receiving unsatisfactory services.

## VII. EXPERIMENT

We validate the effectiveness of the proposed service provision framework through experiments with scenarios. We first describe the experimental setting and the implementation details. Then, we present and analyze the experiment results. We also report the experiment videos through IEEE Xplore.

### A. Environment

We constructed a Smart Home environment for the verification of the effectiveness of the proposed SF-EM and the service provision framework. We set up a kitchen environment using the proposed service framework where robot and IoT systems provide services to the user. We used Mybot developed at the KAIST Robot Intelligence Technology Laboratory. The robotic head connected to the upper body through a neck which has three degree of freedoms (DoFs) is equipped with an RGB-D camera for image processing. We used a color mark for object recognition. Mybot's upper body contains two arms (ten DoFs for each arm) and one trunk (two DoFs). The lower body of Mybot includes an omni-directional wheel-base and a power supply. Mybot can perform home chore tasks [33], [34], gaze control [35], and autonomous conversation [36]. For the implementation of the IoT system, we utilized Raspberry Pi 3, two types of sensors, and Microbot Push [37]. We set up touch sensors between doors to recognize door opening, and Mircrobot Push to turn on and off the light. We used the illumination sensor to measure the brightness of the room. We collected training data through Mongo DB [38], Web hook, and IFTTT services [39]. Web hook and IFTTT services were also utilized in implementing services that were provided by the IoT system. The learning parameters were the same as those used in the second simulation (feedback mechanism simulation).

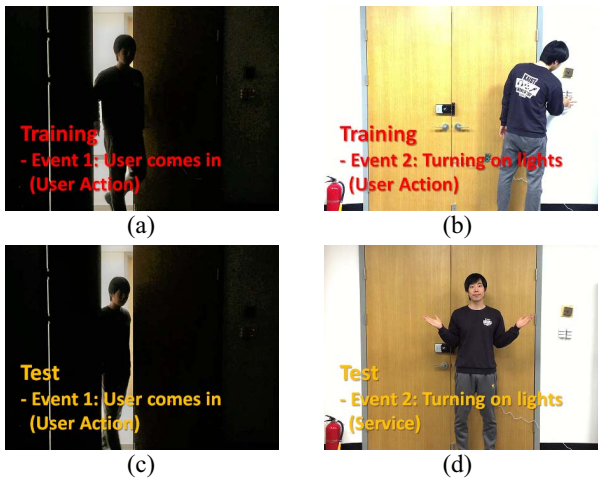

Fig. 10.   Experiment scenario 1. The framework learned the user behavior and offered the service "turning on the light when it is dark."

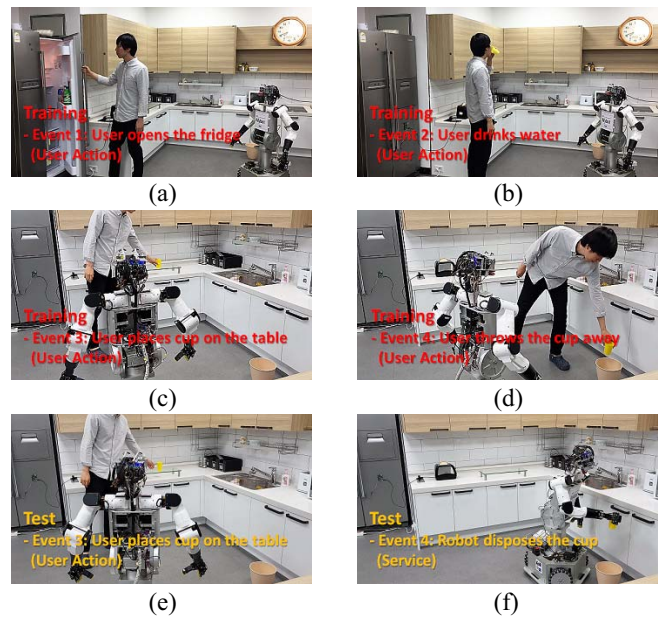

Fig. 11.   Experiment scenario 2. The robot and IoT work together for data collection and service provision in the proposed framework. After the user drank water using a disposal cup, the framework provided the service "disposal of a cup."

### B. Experiment With Scenarios

We evaluated the proposed service provision framework with two scenarios. Each experiment with a scenario consists of training and testing phases. We first trained the system with each scenario and then tested the performance. The first scenario did not entail a comparison study, while we compared the performance of the proposed framework with the conventional memory network in the second scenario.

*1) Scenario 1:* The first scenario is composed of two events, as shown in Fig. 10. In the training phase of the scenario, a user entered the kitchen opening the door when it was dark, which is the first event. Then, the user turned on the light. The touch sensor detected the door opening, and the Microbot Push attached above the light buttons recognized the light turning on. After the memory unit in the proposed framework learned the service episode, the user entered the kitchen again when it was dark. The proposed framework perceived the user entering the kitchen using the IoT system (the touch sensor) and retrieved the first episode successfully. Then, the framework provided the user with the corresponding service, turning on the light. The Microbot Push belonging to the IoT system provided the service rather than the robotic system since it was easier for the IoT system to implement the service.

*2) Scenario 2:* The second scenario is composed of four events as depicted in Fig. 11. The following is the training phase of the second scenario.

1) The user opens the refrigerator.
2) The user drinks water with a disposable cup.
3) The user places the empty cup on the table.
4) The user throws away the used cup in the trash bin.

In the second scenario, we evaluated both the retrieval of the correct episode and the appropriate response to the negative feedback due to the change of user preference. For this, we varied the user preference in the middle of the experiment and compared the reaction of the proposed framework with the reaction of Deep ART.

After the proposed framework learned the second scenario, the user opened the refrigerator and drank water using a disposable cup. Then, the user placed the cup on the table. Next, the robot in the service provision framework disposed of the cup. The proposed framework noticed the event that the user opened the refrigerator, using the IoT system (the touch sensor), and observed the user drinking water, using the robot system (image processing with color marking). Then, the robot provided "the disposal of trash" service in contrast to the first scenario where the IoT system implemented the coherent service since the service provided in the second scenario required mobility. In summary, the robot and the IoT systems performed their respective roles but as one integrated unit in the proposed service provision framework. They collaborated on collecting data and providing services.

In the second testing phase, the user modified the behavioral pattern. The user stopped using a disposable cup instead started to use a plastic cup which is reusable. After drinking water, the user wished to keep the plastic cup rather than throwing it away. We compared the proposed SF-ART-based service provision framework against the conventional Deep ART memory network. Fig. 12 displays the experiment snap shots. After drinking water, the user put the plastic cup on the table. Then, the robot tried to dispose off the cup. The user gave negative feedback and the robot returned the cup back to the original place. The user placed the plastic cup into the sink, and the robot observed the color patches attached to the two objects and recognized the new user behavior from the geometric relation between the two patches. This new observation resulted in learning episode 3 which modifies episode 2 for both the proposed framework and Deep ART. In addition, the proposed framework trained the negative memory with episode 2. When the user positioned the plastic cup on the table again, the robot with the proposed framework moved the cup into the sink, while the robot with Deep ART threw away the plastic cup into the trash bin.

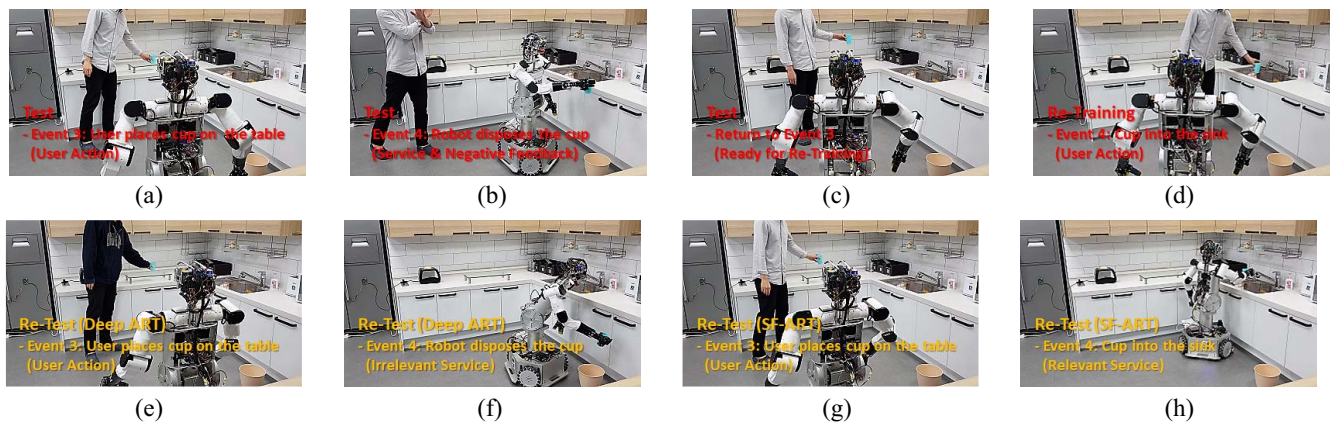

Fig. 12. Second test phase of scenario 2. The user started to use a recyclable cup (change in user preference) and wished to keep the cup after usage. The proposed framework reflected the changed user preference, while the conventional memory network failed to reflect it. (a) Test: event 3. (b) Negative feedback. (c) Returning to event 3. (d) Retraining: event 4. (e) Retest (Deep ART): event 3. (f) Retest (Deep ART): event 4. (g) Retest (SF-ART): event 3. (h) Retest (SF-ART): event 4.

## VIII. Discussion and Conclusion

In this section, we present a few discussion points for the future work and concluding remarks. We noted that we can integrate other systems into the proposed framework to extend the functionality in future work. First, the user can effectively provide the feedback by introducing the emotion recognition system. In this paper, the system operator controlled the software during the experiment to reflect the provided feedback. However, this feedback provision method is unnatural and inconvenient. With the emotion recognition system introduced to the proposed framework, the user can provide both positive and negative feedbacks naturally and comfortably through facial expressions or gestures. Second, the introduction of remote control systems and mobile systems will enable more diverse services for the proposed framework. By introducing such systems, the framework can learn what services the user wants when the user is not in the house. For example, a user wishes to set up the house temperature a few minutes before entering the house or to prepare food in advance through a microwave oven. The framework can learn these services by collecting user information through GPS in the mobile system and remote control function.

For the second point, we can analyze a part of design choices, such as doubling the memory strength and attenuating the vigilance parameter twice upon receiving a positive feedback in the following research. From a series of experiments during the process of formulating the memory parameter modulation, we found that higher factors for positive feedbacks saturate the memory strength in a shorter period of time and diminish the vigilance parameter dramatically. In addition, the effect of feedbacks becomes minimal with lower factors. Then, we determined the factors empirically. However, it is true that we need to conduct mathematical analysis of such design choices in depth and find out the principles behind it. In the future work, we can analyze the effects of design choices mathematically and provide guidelines for setting such factors. Design choices for further analysis include the factors for memory strength modulation, vigilance parameter modulation, and coefficient for code activation in (23).

In the conclusion, we proposed an SF ART, designed the SF-EM based on SF-ART, and a service provision framework for Smart Home. The proposed SF-ART and SF-EM networks enhance the memory stability by employing the adaptive decay factor. In addition, we designed a feedback mechanism for the proposed memory architectures, which allow the memory architectures to accept user feedbacks and to satisfy the user demand accordingly. In the proposed service provision framework, the robot and IoT systems collaborate as one unit to collect data and provide a service. The proposed service provision framework utilizes the SF-EM designed in this paper as a learning and reasoning module. We conducted simulations to verify the enhanced stability of the proposed memory architecture, and the experiment results showed the improved stability and versatility compared to conventional models. Furthermore, the experiments with the robot and the IoT systems demonstrated the effectiveness of the proposed feedback mechanism and the performance of the service provision framework.

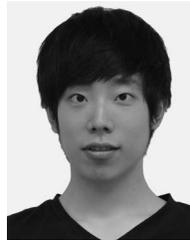

**Ue-Hwan Kim** received the B.S. and M.S. degrees in electrical engineering from the Korea Advanced Institute of Science and Technology, Daejeon, South Korea, in 2015 and 2013, respectively, where he is currently pursuing the Ph.D. degree.

His current research interests include smart home, cognitive IoT, service robot, computational memory systems, and learning algorithms.

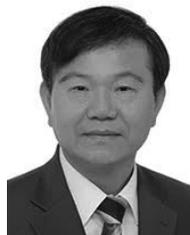

**Jong-Hwan Kim** (F'09) received the Ph.D. degree in electronics engineering from Seoul National University, Seoul, South Korea, in 1987.

Since 1988, he has been with the School of Electrical Engineering, Korea Advanced Institute of Science and Technology (KAIST), Daejeon, South Korea, where he is leading the Robot Intelligence Technology Laboratory as a Professor. He is currently the Dean of the College of Engineering, KAIST, and the Director of the KoYoung-KAIST AI Joint Research Center and the Machine Intelligence and Robotics Multisponsored Research Platform. He has authored five books and five edited books, two journal special issues, and 400 refereed papers in technical journals and conference proceedings. His current research interests include intelligence technology, machine intelligence learning, intelligent interactive technology, ubiquitous and genetic robots, and humanoid robots.

Dr. Kim was an Associate Editor of the IEEE TRANSACTIONS ON EVOLUTIONARY COMPUTATION and the *IEEE Computational Intelligence Magazine*. He has delivered over 200 invited talks on computational intelligence and robotics, including 50 keynote speeches at the international conferences.